\documentclass[11pt,a4paper]{article}
\usepackage[hyperref]{acl2021}
\usepackage{times}
\usepackage{latexsym}

\usepackage{graphicx}
\usepackage{multirow}
\usepackage{xcolor}
\usepackage{booktabs}
\usepackage{amssymb}
\usepackage{array}
\usepackage{amsmath}
\usepackage{stfloats}
\usepackage{adjustbox}
\usepackage{caption}
\usepackage{subcaption}
\usepackage{comment}
\usepackage{mathtools}

\definecolor{goldenpoppy}{rgb}{0.99, 0.76, 0.0}
\definecolor{cadmiumred}{rgb}{0.89, 0.0, 0.13}
\definecolor{ao}{rgb}{0.0, 0.0, 1.0}
\definecolor{ao(english)}{rgb}{0.0, 0.5, 0.0}

\colorlet{NextBlue}{red!00!green!00!blue!100}
\colorlet{NextRed}{red!100!green!00!blue!00}
\colorlet{NextGreen}{red!00!green!75!blue!25}

\newcommand\blfootnote[1]{%
  \begingroup
  \renewcommand\thefootnote{}\footnote{#1}%
  \addtocounter{footnote}{-1}%
  \endgroup
}

\definecolor{purple1}{HTML}{7570b3}
\definecolor{green1}{HTML}{1b9e77}
\definecolor{orange}{HTML}{d95f02}
\definecolor{green2}{HTML}{4daf4a}
\definecolor{red}{HTML}{e41a1c}
\definecolor{blue}{HTML}{377eb8}
\definecolor{purple2}{HTML}{984ea3}

\usepackage{microtype}
\aclfinalcopy 

\title{Learn to Resolve Conversational Dependency: \\
A Consistency Training Framework for Conversational \\ Question Answering}

\author{Gangwoo Kim \quad Hyunjae Kim \quad Jungsoo Park \quad Jaewoo Kang$^\dagger$ \\
    Korea University \\
    \texttt{\{gangwoo\_kim,hyunjae-kim\}@korea.ac.kr} \\ 
    \texttt{\{jungsoo\_park,kangj\}@korea.ac.kr}
    }

\date{}

\begin{document}
\maketitle

\blfootnote{\textsuperscript{$\dagger$} Corresponding author}

\begin{abstract}

One of the main challenges in conversational question answering (CQA) is to resolve the conversational dependency, such as anaphora and ellipsis.
However, existing approaches do not explicitly train QA models on how to resolve the dependency, and thus these models are limited in understanding human dialogues.
In this paper, we propose a novel framework, \textsc{ExCorD} (\textbf{Ex}plicit guidance on how to resolve \textbf{Co}nve\textbf{r}sational \textbf{D}ependency) to enhance the abilities of QA models in comprehending conversational context.
\textsc{ExCorD} first generates self-contained questions that can be understood without the conversation history, then trains a QA model with the pairs of original and self-contained questions using a consistency-based regularizer.
In our experiments, we demonstrate that \textsc{ExCorD} significantly improves the QA models' performance by up to 1.2 F1 on QuAC \cite{choi2018quac}, and 5.2 F1 on CANARD \cite{elgohary2019can}, while addressing the limitations of the existing approaches.\footnote{Our models and code are available at: \\ 
\url{https://github.com/dmis-lab/excord}}

\end{abstract}

\section{Introduction}

\begin{figure} [t]
    \centering
    \includegraphics[width=\columnwidth]{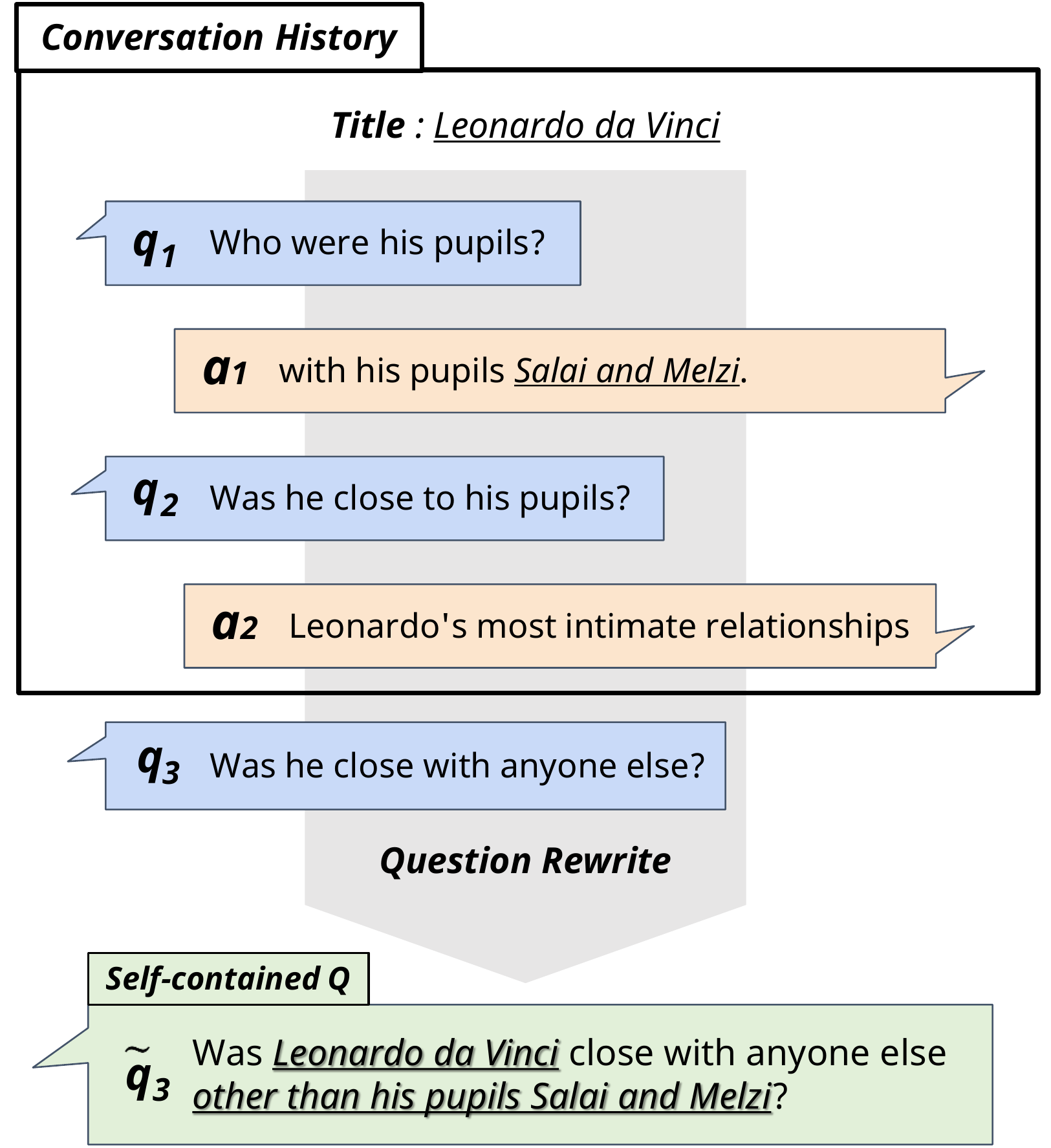}
    \vspace{-0.1cm}\caption{An example of the QuAC dataset \cite{choi2018quac}. 
    Owing to linguistic phenomena in human conversations, such as anaphora and ellipsis, the current question $q_3$ should be understood based on the conversation history: $q_1$, $a_1$, $q_2$, and $a_2$.
    Question $q_3$ can be reformulated as a self-contained question $\tilde{q}_3$ via a question rewriting (QR) process.
    }\vspace{-0.3cm}
    \label{fig:teaser}
    \vspace{-.5cm}
\end{figure}

Conversational question answering (CQA) involves modeling the information-seeking process of humans in a dialogue.
Unlike \textit{single-turn} question answering (QA) tasks \cite{rajpurkar2016squad, kwiatkowski2019natural}, CQA is a \textit{multi-turn} QA task, where questions in a dialogue are context-dependent;\footnote{While the term ``context'' usually refers to the evidence document from which the answer is extracted, in CQA, it refers to conversational context.} hence they need to be understood with the conversation history \cite{choi2018quac, reddy2019coqa}.
As illustrated in Figure \ref{fig:teaser}, to answer the current question ``\emph{Was he close with anyone else?,}'' a model should resolve the conversational dependency, such as anaphora and ellipsis, based on the conversation history.

A line of research in CQA proposes the \textit{end-to-end} approach, where a single QA model jointly encodes the evidence document, the current question, and the whole conversation history \cite{huang2018flowqa, yeh2019flowdelta, qu2019bert}.
In this approach, models are required to automatically learn to resolve conversational dependencies.
However, existing models have limitations to do so without explicit guidance on how to resolve these dependencies. 
In the example presented in Figure \ref{fig:teaser}, models are trained without explicit signals that ``\emph{he}'' refers to ``\emph{Leonardo da Vinci,}''  and ``\emph{anyone else}'' can be more elaborated with ``\emph{other than his pupils, Salai and Melzi}.''

Another line of research proposes a \textit{pipeline} approach that decomposes the CQA task into question rewriting (QR) and QA, to reduce the complexity of the task \cite{vakulenko2020question}.
Based on the conversation history, QR models first generate self-contained questions by rewriting the original questions, such that the self-contained questions can be understood without the conversation history.
For instance, the current question $q_3$ is reformulated as the self-contained question $\tilde{q}_3$ by a QR model  in Figure \ref{fig:teaser}.
After rewriting the question, QA models are asked to answer the self-contained questions rather than the original questions.
In this approach, QA models are trained to answer relatively simple questions whose dependencies have been resolved by QR models. 
Thus, this limits reasoning abilities of QA models for the CQA task, and causes QA models to rely on QR models.

In this paper, we emphasize that QA models can be enhanced by using both types of questions with explicit guidance on how to resolve the conversational dependency.
Accordingly, we propose \textsc{ExCorD} (\textbf{Ex}plicit guidance on how to Resolve \textbf{Co}nve\textbf{r}sational \textbf{D}ependency), a novel training framework for the CQA task.
In this framework, we first generate self-contained questions using QR models.
We then pair the self-contained questions with the original questions, and jointly encode them to train QA models with consistency regularization \cite{laine2016temporal, xie2019unsupervised}.
Specifically, when original questions are given, we encourage QA models to yield similar answers to those when self-contained questions are given.
This training strategy helps QA models to better understand the conversational context, while circumventing the limitations of previous approaches.

To demonstrate the effectiveness of \textsc{ExCorD}, we conduct extensive experiments on the three CQA benchmarks.
In the experiments, our framework significantly outperforms the existing approaches by up to 1.2 F1 on QuAC \cite{choi2018quac} and by 5.2 F1 on CANARD \cite{elgohary2019can}.
In addition, we find that our framework is also effective on a dataset CoQA  \cite{reddy2019coqa} that does not have the self-contained questions generated by human annotators. This indicates that the proposed framework can be adopted on various CQA datasets in future work.
We summarize the contributions of this work as follows:
\begin{itemize}
    \item We identify the limitations of previous approaches and propose a unified framework to address these. Our novel framework improves QA models by incorporating QR models, while reducing the reliance on them.
    
    \item Our framework encourages QA models to learn how to resolve the conversational dependency via consistency regularization. To the best of our knowledge, our work is the first to apply the consistency training framework to the CQA task. 
    \item We demonstrate the effectiveness of our framework on three CQA benchmarks. Our framework is model-agnostic and systematically improves the performance of QA models.
\end{itemize}

\section{Background}

\begin{figure*} [t]
    \centering
    \begin{subfigure}[b]{0.3\textwidth}
    \centering
    \includegraphics[height=0.25\paperheight]{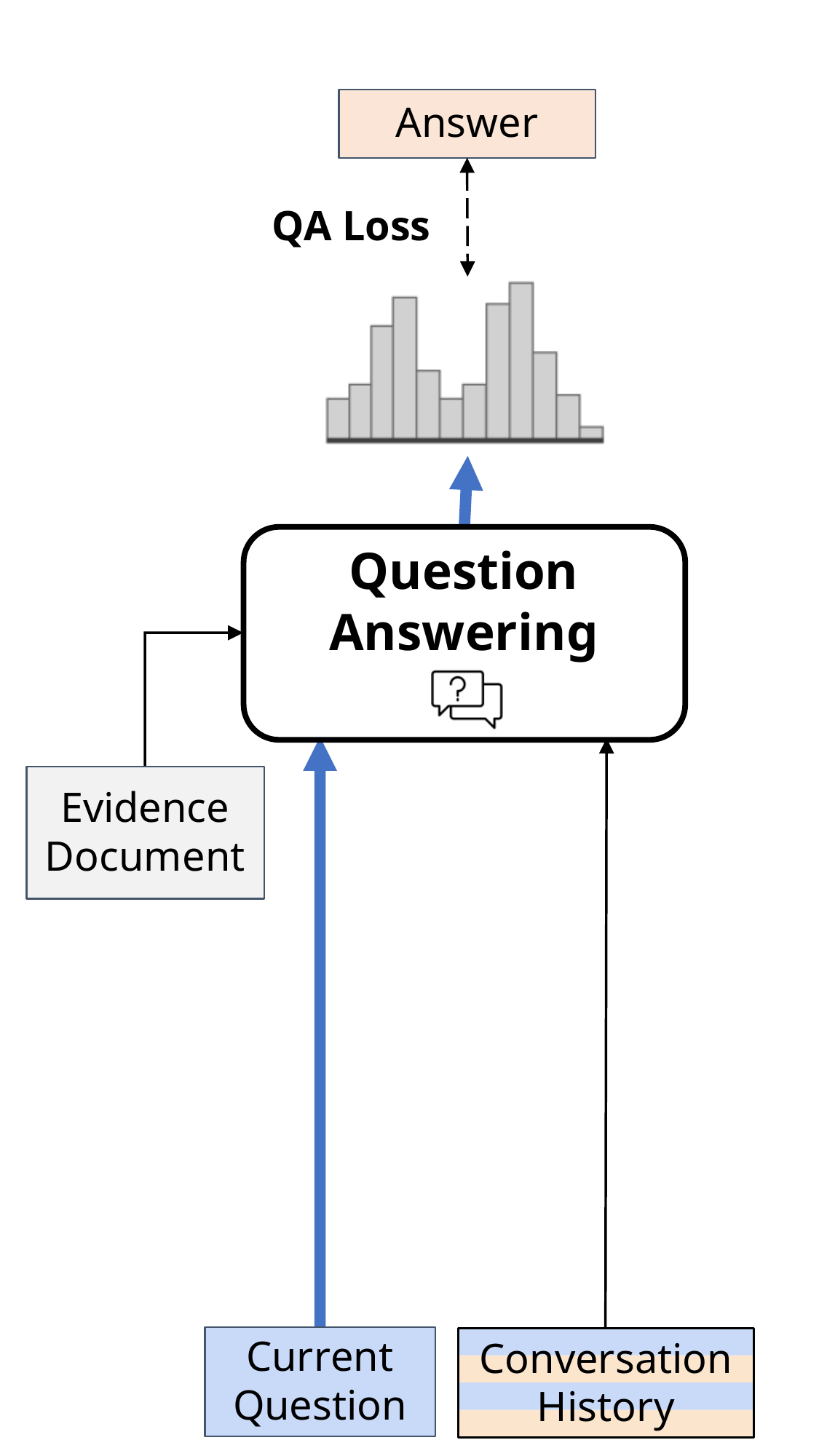}
        \caption{End-to-end approach}
        \label{fig:end2end}
    \end{subfigure}
    \hfill
    \begin{subfigure}[b]{0.3\textwidth}
        \centering
        \includegraphics[height=0.25\paperheight]{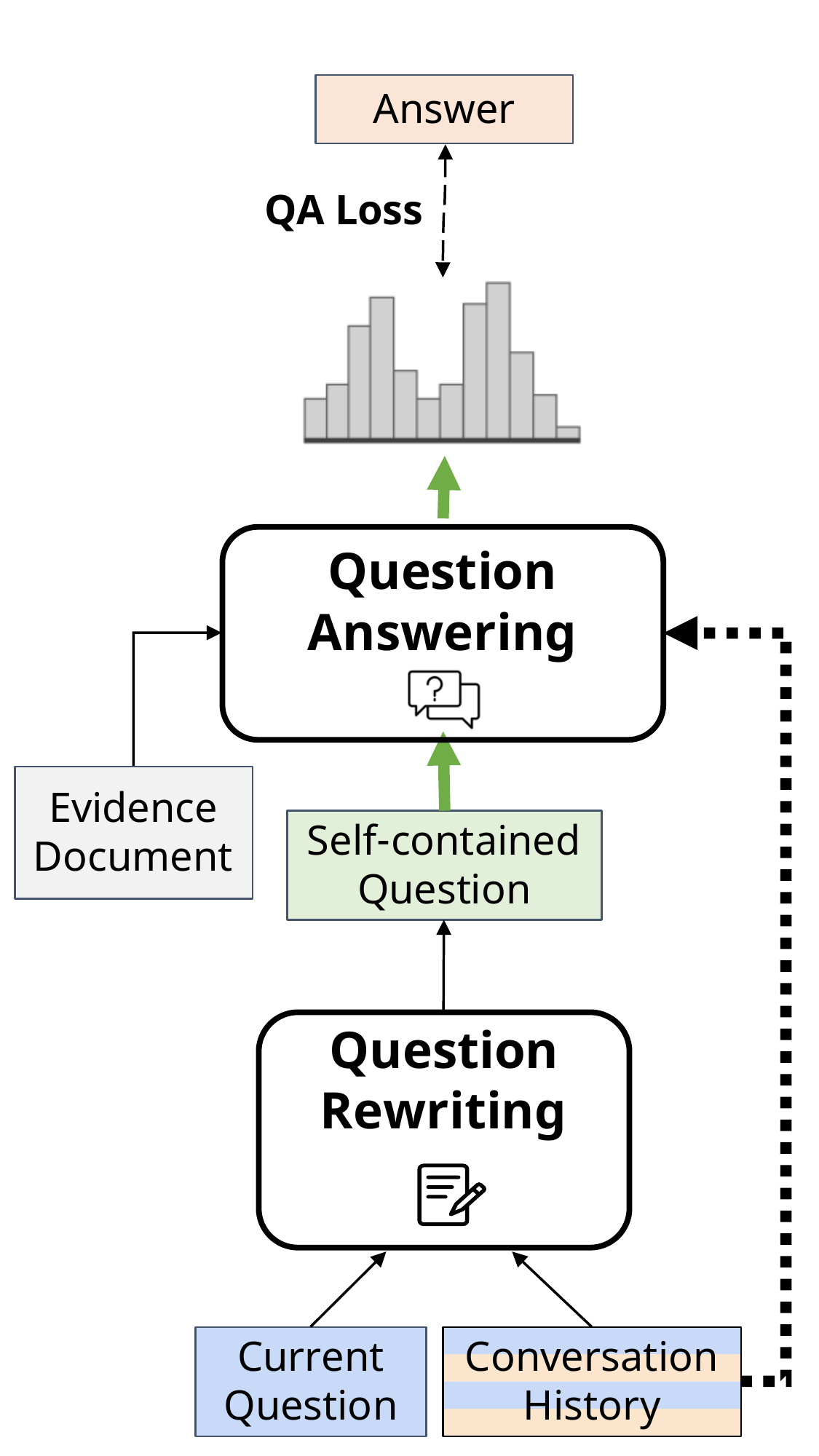}
        \caption{Pipeline approach}
        \label{fig:pipeline}
    \end{subfigure}
    \hfill
    \begin{subfigure}[b]{0.37\textwidth}
        \centering
        \includegraphics[height=0.25\paperheight]{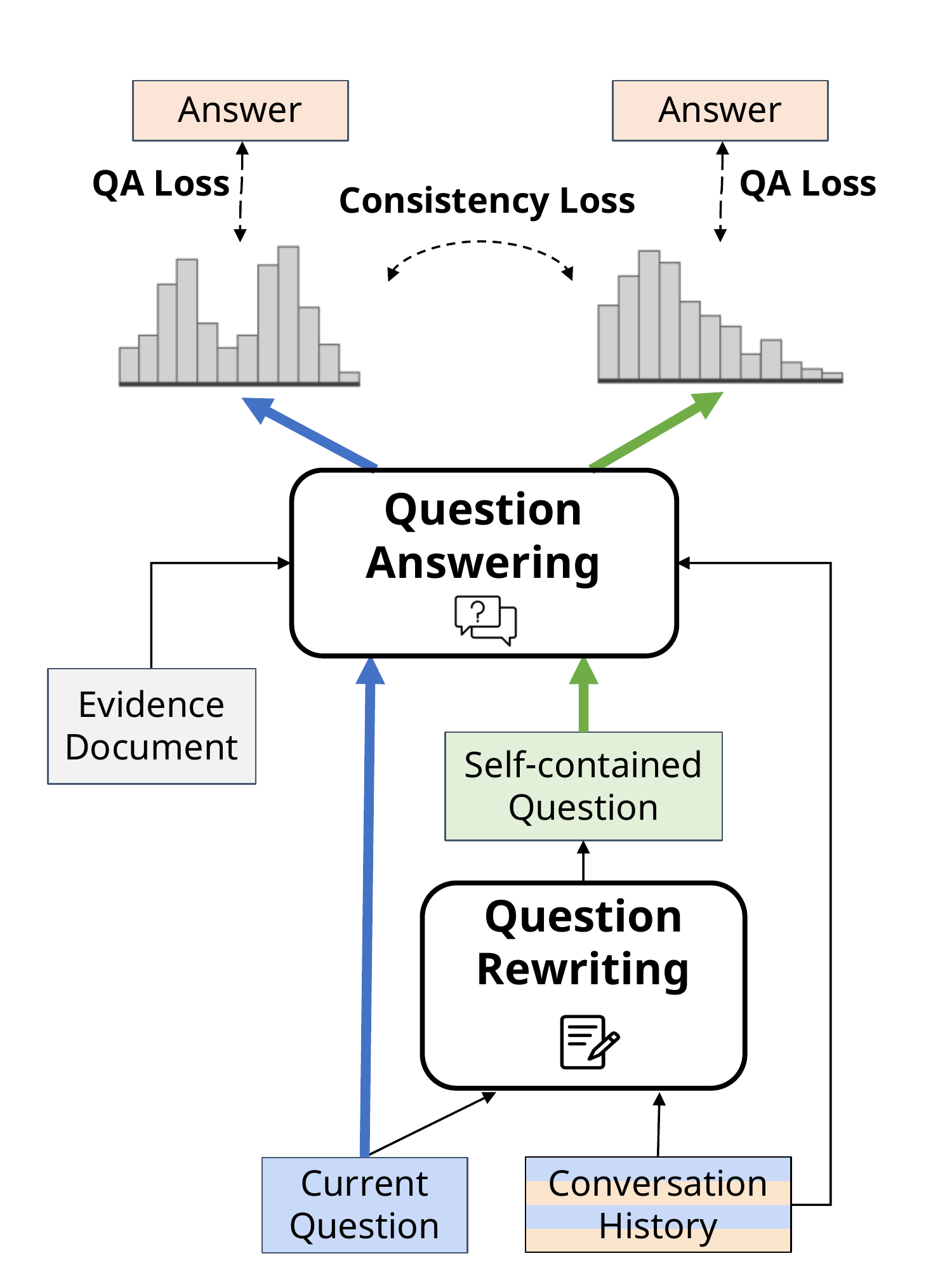}
        \caption{Ours}
        \label{fig:ours}
    \end{subfigure}
    
    \vspace{-0.1cm}
    \caption{Overview of the end-to-end approach, the pipeline approach, and ours.
    In the end-to-end approach, QA models are asked to answer the original questions based on the conversation history.
    In the pipeline approach, the self-contained questions are generated by a QR model, and then QA models answer them.
    Standard QA models are commonly used in this approach; however conversational QA models that encode the history can be adopted (the dotted line in Figure (b)).
    In ours, the original and self-contained question are jointly encoded to train QA models with the consistency loss.}\vspace{-0.3cm}
    \label{fig:overview}
\end{figure*}

\subsection{Task Formulation}
In CQA, a single instance is a dialogue, which consists of an evidence document $d$, a list of questions $\mathbf{q} = [q_1,...,q_{T}]$, and a list of answers for the questions  $\mathbf{a} = [a_1,...,a_T]$, where $T$ represents the number of \textit{turns} in the dialogue.
For the $t$-th turn, the question $q_t$ and the conversation history $\mathcal{H}_{t} = [(q_1,a_1),...,(q_{t-1},a_{t-1})]$ are given, and a model should extract the answer from the evidence document as:

\begin{equation}
    \hat{a}_{t} = \operatorname*{arg\,max}_{a_t} \mathrm{P}(a_t|d, q_t, \mathcal{H}_t)
\end{equation}
where $\mathrm{P}(\cdot)$ represents a likelihood function over all the spans in the evidence document, and $\hat{a}_t$ is the predicted answer.
Unlike single-turn QA, since the current question $q_t$ is dependent on the conversation history $\mathcal{H}_t$, it is important to effectively encode the conversation history and resolve the conversational dependency in CQA.

\subsection{End-to-end Approach}

A naive approach in solving CQA is to train a model in an end-to-end manner (Figure \ref{fig:end2end}).
Since standard QA models generally are ineffective in the CQA task, most studies attempt to develop a QA model structure or mechanism for encoding the conversation history effectively \cite{huang2018flowqa, yeh2019flowdelta, qu2019bert, qu2019attentive}.
Although these efforts improved performance on the CQA benchmarks, existing models remain limited in understanding conversational context.
In this paper, we emphasize that QA models can be further improved with explicit guidance using self-contained questions effectively.

\subsection{Pipeline Approach}

Recent studies decompose the task into two sub-tasks to reduce the complexity of the CQA task.
The first sub-task, question rewriting, involves generating self-contained questions by reformulating the original questions.
Neural-net-based QR models are commonly used to obtain self-contained questions \cite{lin2020conversational, vakulenko2020question}.
The QR models are trained on the CANARD dataset \cite{elgohary2019can}, which consists of 40K pairs of original QuAC questions and their self-contained versions that are generated by human annotators.

After generating the self-contained questions, the next sub-task, question answering, is carried out.
Since it is assumed that the dependencies in the questions have already been resolved by QR models, existing works usually use standard QA models (not specialized to CQA); however conversational QA models can also be used (the dotted line in Figure \ref{fig:pipeline}).
We formulate the process of predicting the answer in the pipeline approach as:

\begin{equation}
\begin{aligned}
   & \mathrm{P}(a_t|d, q_t, \mathcal{H}_t) \approx \\ 
   & \mathrm{P}^\text{rewr}(\tilde{q}_{t}|q_t, \mathcal{H}_t) \cdot \mathrm{P}^\text{read}(a_t|d, \tilde{q}_t)
\end{aligned}
\label{equation:pipeline}
\end{equation}
where $\mathrm{P}^\text{rewr}(\cdot)$ and $\mathrm{P}^\text{read}(\cdot)$ are the likelihood functions of QR and QA models, respectively.
$\tilde{q}_t$ is a self-contained question rewritten by the QR model.

The main limitation of the pipeline approach is that QA models are never trained on the original questions, which limits their abilities to understand the conversational context.  
Moreover, this approach makes QA models dependent on QR models; hence QA models suffer from the error propagation from QR models.
\footnote{We present an example of the error propagation in Section \ref{subsec:error_case}.}
On the other hand, our framework enhances QA models' reasoning abilities for CQA by jointly utilizing original and self-contained questions. 
In addition, QA models in our framework do not rely on QR models at inference time and thus do not suffer from error propagation.

\section{\textsc{ExCorD}: Explicit Guidance on Resolving Conversational Dependency}

We introduce a unified framework that jointly encodes the original and self-contained questions as illustrated in Figure \ref{fig:ours}.
Our framework consists of two stages: (1) generating self-contained questions using a QR model (\textsection\ref{subsec:question_rewriting}) and (2) training a QA model with the original and self-contained questions via consistency regularization (\textsection \ref{subsec:consistency_reg}).

\subsection{Question Rewriting}
\label{subsec:question_rewriting}

Similar to the pipeline approach, we utilize a QR model to obtain self-contained questions.
We use the obtained questions for explicit guidance in the next stage.
As shown in Equation \ref{equation:pipeline}, the QR task is to generate a self-contained question given an original question and a conversation history.
Following \citet{lin2020conversational}, we adopt a T5-based sequence generator \cite{raffel2020exploring} as our QR model, which achieves comparable performance with that of humans in QR.\footnote{On CANARD, our QR model achieved comparable performance with the human performance in preliminary experiments.}
For training and evaluating the QR model, we use the CANARD dataset following previous works on QR \cite{lin2020conversational, vakulenko2020question}.
During inference, we utilize the top-k random sampling decoding based on beam search with the adjustment of the softmax temperature \cite{fan2018hierarchical, xie2019unsupervised}.

\subsection{Consistency Regularization}
\label{subsec:consistency_reg}

Our goal is to enhance the QA model's ability to understand conversational context.
Accordingly, we use consistency regularization \cite{laine2016temporal, xie2019unsupervised}, which enforces a model to make consistent predictions in response to transformations to the inputs.
We encourage the model's predicted answers from the original questions to be similar to those from the self-contained questions (\textsection \ref{subsec:question_rewriting}).
Our consistency loss is defined as:
\begin{equation}
    \mathcal{L}_t^\text{cons} = \text{KL}(\mathrm{P}^\text{read}_{\theta}(a_t|d,q_t,\mathcal{H}_t)||\mathrm{P}^{\text{read}}_{\bar{\theta}}(a_t|d,\tilde{q}_t,\tilde{\mathcal{H}}_t))
\label{equation:consistency_reg}
\end{equation}
where $\text{KL}(\cdot)$ represents the Kullback–Leibler divergence function between two probability distributions. $\theta$ is the model's parameters, and $\bar{\theta}$ depicts a fixed copy of $\theta$.

With the consistency loss, QA models are regularized to make consistent predictions, regardless of whether the given question is self-contained or not.
In order to output an answer distribution that is closer to $\mathrm{P}^{\text{read}}_{\bar{\theta}}(a_t|d,\tilde{q}_t,\tilde{\mathcal{H}}_t)$, QA models should treat original questions as if they were rewritten into self-contained questions by referring to the conversation history.
Through this process, our consistency regularization method serves as explicit guidance that encourages QA models to resolve the conversational dependency.
In our framework, $\mathrm{P}^\text{read}_{\theta}(a_t|\cdot)$ is the answer span distribution over all evidence document tokens.
In contrast to \citet{asai2020logic}, by using all probability values in the answer distributions, the signals of self-contained questions can be effectively propagated to the QA model.
In addition to using all probability values, we also sharpened the target distribution $\mathrm{P}^{\text{read}}_{\bar{\theta}}(a_t|d,\tilde{q}_t,\tilde{\mathcal{H}}_t)$ by adjusting the temperature \citep{xie2019unsupervised} to strengthen the QA model's training signal.

Finally, we calculate the final loss as:
\begin{equation}
 \mathcal{L} =  \mathcal{L}^\text{orig} + \lambda_1 \mathcal{L}^\text{self} + \lambda_2 \mathcal{L}^\text{cons}
\label{equation:total_loss}
\end{equation}
\noindent where $\lambda_1$ and $\lambda_2$ are hyperparameters.
$\mathcal{L}^{\text{orig}}$ and $\mathcal{L}^{\text{self}}$ are calculated by the negative log-likelihood between the predicted answers and gold standards given the original and self-contained questions, respectively.

\paragraph{Comparison with previous works}
Consistency training has mainly been studied as a method for regularizing model predictions to be invariant to small noises that are injected into the input samples \cite{sajjadi2016regularization, laine2016temporal, miyato2016adversarial, xie2019unsupervised}.
The intuition behind consistency training is to push noisy inputs closer towards their original versions.
Therefore, only the original parameters (i.e., $\theta$) are updated, while the copied model parameters (i.e., $\bar{\theta}$) are fixed.

In contrast to the original concept of consistency training, our goal is to go in the opposite direction and update the original parameters.
Thus, we fix the parameters $\bar{\theta}$ with self-contained questions, and soley update $\theta$ for each training step as shown in Equation \ref{equation:consistency_reg}.

\section{Experiments}
In this section, we describe our experimental setup and compare our framework to baseline approaches (i.e., the end-to-end and pipeline approaches).

\subsection{Datasets}

\paragraph{QuAC}
QuAC \cite{choi2018quac} comprises 100k QA pairs in information-seeking dialogues, where a student asks questions based on a topic with background information provided, and a teacher provides the answers in the form of text spans in Wikipedia documents.
Since the test set is only available in the QuAC challenge, we evaluate models on the development set.\footnote{\url{https://quac.ai/}}
For validation, we use a subset of the original training set of QuAC, which consists of questions that correspond to the self-contained questions in CANARD's development set.
The remaining data is used for training.

\paragraph{CANARD}
CANARD \cite{elgohary2019can} consists of 31K, 3K, and 5K QA pairs for training, development, and test sets, respectively.
The questions in CANARD are generated by rewriting a subset of the original questions in QuAC.
We use the training and development sets for training and validating QR models, and the test set for evaluating QA models.

\paragraph{CoQA}
CoQA \cite{reddy2019coqa} consists of 127K QA pairs and evidence documents in seven domains.
In terms of the question distribution, CoQA significantly differs from QuAC (see \textsection \ref{subsec:transfer}).
We use CoQA to test the transferability of \textsc{ExCorD}, where a QR model trained on CANARD generates the self-contained questions in a zero-shot manner. 
Subsequently, we train a QA model by using the original and synthetic questions.
Similar to QuAC, the test set of CoQA is soley available in the CoQA challenge. \footnote{\url{https://stanfordnlp.github.io/coqa/}}
Therefore, we randomly sample 5\% of the QA dialogues in the training set and adopt them as our development set.

\subsection{Metrics}
Following \citet{choi2018quac}, we use the F1, HEQ-Q, and HEQ-D for QuAC and CANARD. 
HEQ-Q measures whether a model finds more accurate answers than humans (or the same answers) in a given question.
HEQ-D measures the same thing, but in a given dialog instead of a question.
For CoQA, we report the F1 scores for each domain (children's story, literature from Project Gutenberg, middle and high school English exams, news articles from CNN, Wikipedia) and the overall F1 score, as suggested by \citet{reddy2019coqa}.

\subsection{QA models}
Note that the baseline approaches and our framework do not limit the structure of QA models.
For a fair comparison of the baseline approaches and \textsc{ExCorD}, we test the same QA models in all approaches.
The selected QA models are commonly used and have been proven to be effective in CQA.

\paragraph{BERT}
BERT \cite{devlin2019bert} is a contextualized word representation model that is pretrained on large corpora.
BERT also works well on CQA datasets, although it is not designed for CQA.
It receives the evidence document, current question, and conversation history of the previous turn as input.

\paragraph{BERT+HAE}
BERT+HAE is a BERT-based QA model with a CQA-specific module.
Following \citet{qu2019bert}, we add the history answer embedding (HAE) to BERT's word embeddings.
HAE encodes the information of the answer spans from the previous questions.

\paragraph{RoBERTa}
RoBERTa \cite{liu2019roberta} improves BERT by using pretraining techniques to obtain the robustly optimized weights on larger corpora.
In our experiments, we found that RoBERTa performs well in CQA, achieving comparable performance with the previous SOTA model, HAM \cite{qu2019attentive}, on QuAC.
Thus, we adopt RoBERTa as our main baseline model owing to its simplicity and effectiveness.
It receives the same input as BERT, otherwise specified.

\subsection{Implementation Details}

The CANARD training set provides 31,527 self-contained questions from the original QuAC questions.
Therefore, we can obtain 31,527 pairs of original and self-contained questions without question rewriting.
For the rest of the original questions, we automatically generate self-contained questions by using our QR model.
Finally, we obtain 83,568 question pairs and use them in our consistency training.
We denote the original questions, self-contained questions generated by humans, and self-contained questions generated by a QR model as $\mathbf{Q}$, $\tilde{\mathbf{Q}}^{\text{human}}$, and $\tilde{\mathbf{Q}}^{\text{syn}}$, respectively.
Additional implementation details are described in Appendix  \ref{subsec:hyperparameters}

\begin{table*}[t]
\footnotesize
\centering
\begin{tabular}{ll|lll|lll}
\toprule
\multirow{2}{*}{QA Model} & \multirow{2}{*}{Approach} & \multicolumn{3}{c}{QuAC} & \multicolumn{3}{c}{CANARD} \\
& & \multicolumn{1}{c}{F1} & \multicolumn{1}{c}{HEQ-Q} & \multicolumn{1}{c}{HEQ-D} & \multicolumn{1}{c}{F1} & \multicolumn{1}{c}{HEQ-Q} & \multicolumn{1}{c}{HEQ-D}  \\
\midrule
\multirow{3}{*}{BERT} & End-to-end & 61.5 & 57.1 & 5.0 & 57.4 & 52.9 & 3.2 \\
& Pipeline & 61.2 \textcolor{cadmiumred}{(- 0.3)} & 56.8 \textcolor{cadmiumred}{(- 0.3)} & 5.0 (--) & 62.2~\textcolor{ao}{(+ 4.8)}  & 57.8~\textcolor{ao}{(+ 4.9)} & 6.0~\textcolor{ao}{(+ 2.8)} \\
& Ours & \textbf{62.7}~\textcolor{ao}{(+ 1.2)} & \textbf{58.4}~\textcolor{ao}{(+ 1.3)} & \textbf{6.0}~\textcolor{ao}{(+ 1.0)} & \textbf{62.6}~\textcolor{ao}{(+ 5.2)} & \textbf{58.2}~\textcolor{ao}{(+ 5.3)} & \textbf{6.4}~\textcolor{ao}{(+ 3.2)} \\ \midrule 
\multirow{3}{*}{BERT+HAE} & End-to-end & 62.0 & 57.3 & 5.5 & 58.2 & 53.5 & 5.5 \\
& Pipeline & 61.1 \textcolor{cadmiumred}{(- 0.9)} & 56.3 \textcolor{cadmiumred}{(- 1.0)} & 5.0 \textcolor{cadmiumred}{(- 0.5)} & 62.4~\textcolor{ao}{(+ 4.2)} & 57.8~\textcolor{ao}{(+ 4.3)} & \textbf{6.0}~\textcolor{ao}{(+ 0.5)} \\
& Ours & \textbf{63.2}~\textcolor{ao}{(+ 1.2)} & \textbf{58.9}~\textcolor{ao}{(+ 1.6)} & \textbf{5.7}~\textcolor{ao}{(+ 0.2)} & \textbf{63.1}~\textcolor{ao}{(+ 4.9)} & \textbf{58.4}~\textcolor{ao}{(+ 4.9)} & 5.7~\textcolor{ao}{(+ 0.2)} \\ \midrule 
\multirow{3}{*}{RoBERTa} & End-to-end & 66.5 & 62.4 & 7.2 & 65.8 & 62.2 & 7.1 \\
& Pipeline & 65.2 \textcolor{cadmiumred}{(- 1.3)} & 60.9 \textcolor{cadmiumred}{(- 1.5)} & 7.1 \textcolor{cadmiumred}{(- 0.1)} & 66.9~\textcolor{ao}{(+ 1.1)} & 63.2~\textcolor{ao}{(+ 1.0)} & 7.3~\textcolor{ao}{(+ 0.2)} \\
& Ours & \textbf{67.7}~\textcolor{ao}{(+ 1.2)} & \textbf{64.0}~\textcolor{ao}{(+ 1.6)} & \textbf{9.3}~\textcolor{ao}{(+ 2.1)} & \textbf{68.1}~\textcolor{ao}{(+ 2.3)} & \textbf{64.2}~\textcolor{ao}{(+ 2.0)} & \textbf{8.4}~\textcolor{ao}{(+ 1.3)} \\ \bottomrule
\end{tabular}
\caption{Comparison in performance of the baseline approaches and our framework on QuAC and CANARD.
The best scores are highlighted in bold.}\vspace{-0.3cm}
\label{table:main_results}
\end{table*}

\subsection{Results}
\label{subsec:results}
Table \ref{table:main_results} presents the performance comparison of the baseline approaches to our framework on QuAC and CANARD.
Compared to the end-to-end approach, \textsc{ExCorD} consistently improves the performance of QA models on both datasets. 
Also, these improvements are significant: \textsc{ExCorD} improves the performance of the RoBERTa by absolutely 1.2 and 2.3 F1 scores and BERT by 1.2 and 5.2 F1 scores on QuAC and CANARD, respectively.
From these results, we conclude that the consistency training with original and self-contained questions enhances ability of QA models to understand the conversational context.

On QuAC, the pipeline approach underperforms the end-to-end approach in all baseline models.
This indicates that training a QA model soley with self-contained questions is ineffective  when human rewrites are not given at the inference phase.
On the other hand, \textsc{ExCorD} improves QA models by using both types of questions.
As presented in Table \ref{table:main_results}, our framework significantly outperforms the baseline approaches on QuAC.

On CANARD, the pipeline approach is significantly more effective than the end-to-end approach. Since QA models are trained with self-contained questions in the pipeline approach, they perform well on CANARD questions.
Nevertheless, \textsc{ExCorD} still outperforms the pipeline approach in most cases.
Compared to the pipeline approach, our framework improves the performance of RoBERTa by absolutely 1.2 F1 score.

\section{Analysis and Discussion}

We elaborate on analyses regarding component ablation and transferability. We also describe a case study carried out to highlight such differences between our and baseline approaches.

\subsection{Ablation Study}
\label{subsec:ablation}

In this section, we comprehensively explore the factors contributing to this improvement in detail:
(1) using self-contained questions that are rewritten by humans ($\tilde{\mathbf{Q}}^{\text{human}}$) as additional data, (2) using self-contained questions that are synthetically generated by the QR model ($\tilde{\mathbf{Q}}^{\text{syn}}$), and (3) training a QA model with our consistency framework.
In Table \ref{table:ablation}, we present the performance gaps when each component is removed from our framework.
We use RoBERTa on QuAC in this experiment.

We first explore the effects of $\tilde{\mathbf{Q}}^{\text{human}}$ and $\tilde{\mathbf{Q}}^{\text{syn}}$.
As shown in Table \ref{table:ablation}, excluding $\tilde{\mathbf{Q}}^{\text{human}}$ degrades the performance of RoBERTa in our framework.
Although automatically generated, $\tilde{\mathbf{Q}}^{\text{syn}}$ contributes to the performance improvement.
Therefore, both types of self-contained questions are useful in our framework.

To investigate the effect of our framework, we simply augment $\tilde{\mathbf{Q}}^{\text{human}}$ and $\tilde{\mathbf{Q}}^{\text{syn}}$ to $\mathbf{Q}^{\text{orig}}$, which is called Question Augment (question data augmentation).
We find that Question Augment slightly improves the performance of RoBERTa on CANARD, whereas it degrades the performance on QuAC.
This shows that simply augmenting the questions is ineffective and does not guarantee improvement.
On the other hand, our consistency training approach significantly improves performance, showing that \textsc{ExCorD} is a more optimal way to utilizing self-contained questions.

\begin{table}[t!]
\footnotesize
\centering
\resizebox{0.48\textwidth}{!}{
\begin{tabular}{l|cc}
\toprule
 \multirow{2}{*}{Method} & QuAC & CANARD \\
& F1 & F1 \\
 \midrule
 \textsc{ExCorD}  & 67.7 & 68.1 \\ 
 \quad -- $\tilde{\mathbf{Q}}^{\text{syn}}$ & 67.5 & 67.7 \\
 \quad -- $\tilde{\mathbf{Q}}^{\text{human}}$ & 67.3 & 67.2 \\ \cmidrule{1-3}
 Question Augment. (w/o. \textsc{ExCorD}) & 65.9 & 66.2 \\
  \quad -- $\tilde{\mathbf{Q}}^{\text{syn}}$ & 66.1& 66.5 \\
  \quad -- $\tilde{\mathbf{Q}}^{\text{human}}$ & 65.3 & 66.0 \\
 \quad -- $\tilde{\mathbf{Q}}^{\text{syn}}$, $\tilde{\mathbf{Q}}^{\text{human}}$ (End-to-end)  & 66.5 & 65.8 \\
\bottomrule
\end{tabular}}
\caption{Effect of self-contained questions and our consistency framework. We use RoBERTa in this experiment.}\vspace{-0.3cm}
\label{table:ablation}
\end{table}

\begin{table*}[t]
\centering
\resizebox{\textwidth}{!}{
\begin{tabular}{l}
\toprule
\textbf{Error case \# 1} \\
\textbf{Title} : Montgomery Clift \quad \textbf{Section Title} : Film career \\ 
\textbf{Document $d$} : \\
$\cdots$ \\
\colorbox{red!40}{His second movie} was \colorbox{red!40}{The Search}. Clift was unhappy with the quality of the script, and edited it himself. \colorbox{red!40}{The movie} 
\\ was awarded a screenwriting Academy Award for the credited writers. \\
$\cdots$
\\ \midrule
\textbf{$q_\text{1}$} : When did Clift start his film career? \\
\textbf{$a_\text{1}$} : \colorbox{blue!40}{His first movie} role was opposite John Wayne in \colorbox{blue!40}{Red River}, which was shot in 1946 and released in 1948. \\
\midrule
\textbf{Current Question $q_\text{2}$} : Was \colorbox{blue!40}{the film} a success? \\
\textbf{Human Rewrite $r_\text{2}$} : Was \colorbox{blue!40}{Montgomery Clift's film Red River} a success? \\
\textbf{Golden Answer} : CANNOTANSWER \\
\textbf{Prediction of \emph{End-to-End}} : \colorbox{red!40}{The movie} was awarded a screenwriting Academy Award for the credited writers.
 \\
\textbf{Prediction of \emph{Ours}}  : CANNOTANSWER \\ \midrule\midrule
\textbf{Error case \# 2} \\ 
\textbf{Title} : Train (band) \quad \textbf{Section Title} : 2003-2004: My Private Nation \\ \midrule
$\cdots$ \\
\textbf{$q_\text{5}$} : Did my private nation do any other features? \\
\textbf{$a_\text{5}$} : CANNOTANSWER \\ \midrule
\textbf{Current Question $q_\text{6}$} : Did \underline{my} private nation have any good singles? \\
\textbf{Generated Question $\tilde{q}_\text{6}$} : Did \underline{Train's} private nation have any good singles? \\
\textbf{Golden Answer} : ``Get to Me" (written by Rob Hotchkiss and Pat Monahan) reached number nine on the Billboard \\ Adult Top 40. \\
\textbf{Prediction of \emph{Pipeline}} : CANNOTANSWER \\
\textbf{Prediction of \emph{Ours}}  : ``Get to Me'' (written by Rob Hotchkiss and Pat Monahan) reached number nine on the Billboard \\
Adult Top 40. \\ \bottomrule
\end{tabular}}
\caption{Error analysis for predictions of RoBERTa that are trained with the baseline approaches and \textsc{ExCorD}.
In the first case, the QA model trained with the end-to-end approach fails to resolve the conversational dependency.
The QR model in the second case misunderstands the "my," and generates an unnatural question, triggering an incorrect prediction.
}
\label{table:case_study}
\vspace{-.4cm}
\end{table*}
\subsection{Case Study}
\label{subsec:error_case}

We analyze several cases that the baseline approaches answered incorrectly, but our framework answered correctly.
We also explore how our framework improves the reasoning ability of QA models, compared to the baseline approaches.
These cases are obtained from the development set of QuAC.

The first case in Table \ref{table:case_study} shows the predictions of the two RoBERTa models trained in the end-to-end approach and our framework, respectively.
Note that ``the film'' in the current question does not refer to ``The Search'' (red box) in the document $d$, but ``Red River'' (blue box) in $a_1$.
When trained in the end-to-end approach, the model failed to comprehend the conversational context and misunderstood what ``the film'' refers to, resulting in an incorrect prediction.
On the other hand, when trained in \textsc{ExCorD}, the model predicted the correct answer because it enhances the ability to resolve conversational dependency.

In the second case, we compare the pipeline approach to \textsc{ExCorD}.
In this case, the QR model misunderstood ``my'' in the current question as a pronoun and replaced it with the band's name, ``Train's.''
Consequently, the QA model received the erroneous self-contained question, resulting in an incorrect prediction.
On the other hand, the QA model trained in our framework predicted the correct answer based on the original question $q_6$.

\subsection{Transferability}
\label{subsec:transfer}

We train a QR model to rewrite QuAC questions into CANARD questions.
Then, self-contained questions can be generated for the samples that do not have human rewrites.
This results in the improvement of QA models' performance on QuAC and CANARD (\textsection\ref{subsec:results}).
However, it is questionable whether the QR model can successfully rewrite questions when the original questions significantly differ from those in QuAC.
To answer this, we test our framework on another CQA dataset, CoQA.
We first analyze how the question distributions of QuAC and CoQA differ.
We found that question types in QuAC and CoQA are significantly different, such that QR models could suffer from the gap of question distributions between two datasets.
(See details in Appendix \ref{subsec:comparison_question}).

\begin{table}[t!]
\centering
\resizebox{0.48\textwidth}{!}{
\begin{tabular}{lcccccc}
\toprule
\multirow{2}{*}{QA model} & \multicolumn{5}{c}{CoQA (F1)} \\
& Overall & Child. & Liter. & M\&H & News & Wiki. \\
\midrule
\textbf{BERT} & & & & & &         \\
~~ End-to-End & 78.3 & 77.9 & 73.9 & \textbf{76.4} & 80.6 & 82.7        \\
~~ Pipeline & 76.1 & 75.7 & 73.2 & 74.1 & 78.0 & 79.6        \\
~~ Ours & \textbf{78.8} & \textbf{78.2} & \textbf{75.8} & 75.5 & \textbf{81.3} & \textbf{83.2} \\ \midrule
\textbf{RoBERTa} & & & & & &         \\
~~ End-to-End & 82.8 & 82.5 & 80.2 & \textbf{80.1} & 84.3 & \textbf{87.0}   \\
~~ Pipeline & 81.1 & 81.9 & 78.2 & 78.3 & 82.4 & 85.2        \\
~~ Ours & \textbf{83.4} & \textbf{84.4} & \textbf{81.2} & 79.8 & \textbf{84.6} & \textbf{87.0} \\
\bottomrule
\end{tabular}}
\caption{Effect of our framework on the CoQA dataset that do not have human rewrites. We exclude BERT+HAE for simplification in this experiment.}\vspace{-0.3cm}
\label{table:coqa_results}
\end{table}

To test the transferability of \textsc{ExCorD}, we compare the end-to-end approach to our framework on the CoQA dataset.
Using a QR model trained on CANARD, we generate the self-contained questions for CoQA and train QA models with our framework.
As presented in Table \ref{table:coqa_results}, our framework performs well on CoQA.
The improvement in BERT is 0.5 based on the overall F1, and the performance of RoBERTa is also improved by an overall F1 of 0.6.
Improvements are also consistent in most of the documents' domains.
Therefore, we conclude that our framework can be simply extended to other datasets and improve QA performance even when question distributions are significantly different.
We plan to improve the transferability of our framework by fine-tuning QR models on target datasets in future work.

\section{Related Work}
\paragraph{Conversational Question Answering}

Recently, several works introduced CQA datasets such as \textsc{Quac} \cite{choi2018quac} and \textsc{Coqa} \cite{reddy2019coqa}.
We classified proposed methods to solve the datasets into two approaches: (1) end-to-end and (2) pipeline.
Most works based on the end-to-end approach focused on developing a model structure \cite{zhu2018sdnet, ohsugi2019simple, qu2019bert, qu2019attentive} or training strategy such as multi-task with rationale tagging \cite{ju2019technical} that are specialized in the CQA task or datasets.
Several works demonstrated the effectiveness of the flow mechanism in CQA \cite{huang2018flowqa, chen2019graphflow, yeh2019flowdelta}.

With the advent of a dataset consisting of self-contained questions rewritten by human annotators \cite{elgohary2019can}, the pipeline approach has drawn attention as a promising method for CQA in recent days \cite{vakulenko2020question}.
The approach is particularly useful for the open-domain CQA or passage-retrieval (PR) tasks \cite{dalton2019cast, ren2020conversations, anantha2020open, qu2020open} since self-contained questions can be fed into existing non-conversational search engines such as BM25.
Note that our framework can be used jointly with the pipeline approach in the open-domain setting because our framework can improve QA models' ability to find the answers from the retrieved documents.
We will test our framework in the open-domain setting in future work.

\paragraph{Question Rewriting}

QR has been studied for augmenting training data \cite{buck2018ask, sun2018improving, zhu2019learning, liu2020tell} or clarifying ambiguous questions \cite{min2020ambigqa}.
In CQA, QR can be viewed as a task of simplifying difficult questions that include anaphora and ellipsis in a conversation.
\citet{elgohary2019can} first proposed the question rewriting task as a sub-task of CQA and the \textsc{Canard} dataset for the task, which consists of pairs of original and self-contained questions that are generated by human annotators.
\citet{vakulenko2020question} used a coreference-based model \cite{lee2018higher} and GPT-2 \cite{radford2019language} as QR models and tested the models in the QR and PR tasks.
\citet{lin2020conversational} conducted the QR task using T5 \cite{raffel2020exploring} and achieved on performance comparable to humans on CANARD.
Following \citet{lin2020conversational}, we use T5 in our experiments to generate high-quality questions for enhancing QA models.

\paragraph{Consistency Training}

Consistency regularization~\cite{laine2016temporal, sajjadi2016regularization} has been mainly explored in the context of semi-supervised learning~(SSL)~\cite{chapelle2009semi, oliver2018realistic}, which has been adopted in the textual domain as well ~\cite{miyato2016adversarial, clark2018semi, xie2020unsupervised}.
However, the consistency training framework is also applicable when only the labeled samples are available ~\cite{miyato2018virtual, jiang2019smart, asai2020logic}.
The consistency regularization requires adding noise to the sample, which can be either discrete ~\cite{xie2020unsupervised, asai2020logic, park2021consistency} or continuous~\cite{miyato2016adversarial,jiang2019smart}.
Existing works regularize the predictions of the perturbed samples to be equivalent to be that of the originals'. 
On the other hand, our method encourages the models' predictions for the original answers to be similar to those from the rewritten questions, i.e., synthetic ones.






\section{Conclusion}

We propose a consistency training framework for conversational question answering, which enhances QA models' abilities to understand conversational context.
Our framework leverages both the original and self-contained questions for explicit guidance on how to resolve conversational dependency.
In our experiments, we demonstrate that our framework significantly improves the QA model's performance on QuAC and CANARD, compared to the existing approaches.
In addition, we verified that our framework can be extended to CoQA.
In future work, the transferability of our framework can be further improved by fine-tuning the QR model on target datasets.
Furthermore, future work would include applying our framework to the open-domain setting. 

\section*{Acknowledgements}
We thank Sean S. Yi, Miyoung Ko, and Jinhyuk Lee for providing valuable comments and feedback. This research was supported by the MSIT (Ministry of Science and ICT), Korea, under the ICT Creative Consilience program (IITP-2021-2020-0-01819) supervised by the IITP (Institute for Information \& communications Technology Planning \& Evaluation). This research was also supported by National Research Foundation of Korea (NRF-2020R1A2C3010638).



\bibliographystyle{acl_natbib}
\bibliography{acl2021}

\clearpage

\appendix

\begin{table*}[t]
\centering
\resizebox{\textwidth}{!}{
\begin{tabular}{l|l}
\toprule
QuAC &
CoQA \\ \midrule
\textbf{Title} : \colorbox{blue!40}{Scott Walker} (politician) &
$q_\text{1}$ : Is the US dollar on a decimal system? \\
\textbf{Section Title} : Education &
$a_\text{1}$ : U.S. dollar is based upon a decimal system of values. I \\
$q_\text{1}$ : What kind of education did Scott Walker have? &
$q_\text{2}$ : What country's dollar is not? \\
$a_\text{1}$ : CANNOTANSWER &
$a_\text{2}$ : Unlike the Spanish milled dollar the U.S. dollar is  \\ 
$q_\text{2}$ : Are there any other interesting aspects about this article? & 
based upon a decimal system of values. \\
 
$a_\text{2}$ : signed \colorbox{green!40}{a law to fund evaluation} of the reading skills & 
$q_\text{3}$ : \colorbox{yellow!40}{What is} a mill? \\
of kindergartners as part of an initiative to ensure that students  & 
$a_\text{3}$ : n addition to the dollar the coinage act officially   \\
are reading at or above grade level &
established monetary units of mill or one-thousandth of a dollar \\
\midrule
\textbf{Current Question }$q_\text{3}$ : What other programs did he sign? & 
\textbf{Current Question }$q_\text{4}$ : And a cent? \\
\textbf{Self-contained Question }$\tilde{q}_\text{3}$ :What other programs did \colorbox{blue!40}{Scott Walker}  &
\textbf{Self-contained Question }$\tilde{q}_\text{4}$ : \colorbox{yellow!40}{What is} a cent? \\
sign other than \colorbox{green!40}{a law to fund evaluation} & - \\\bottomrule
\end{tabular}}
\caption{Comparison of questions in QuAC and CoQA. 
In the left side, we can observe several question types that are frequently used in QuAC: unanswerable question ($q_1$) and ``Anything else?'' question ($q_2$). The current question $q_3$ refers to the previous answer (green box) and the background information (blue box).
On the other hand, in the right side, the current question $q_4$ omits the question word that are used in the previous question (yellow box).
} \vspace{-0.3cm}
\label{table:comparison}
\end{table*}

\begin{table}[h]
\footnotesize
\centering
\resizebox{0.35\textwidth}{!}{
\begin{tabular}{l|cc}
\toprule
 Question Type & QuAC & CoQA \\
 \midrule
 Non-factoid & 54 \% & 38~\% \\
 Anything else? & 11 \% &  3$^\dagger$ \% \\
 Unanswerable & 20 \% & 1 \% \\
\bottomrule
\end{tabular}}
\caption{Statistics of question types for QuAC and CoQA. 
All values can be found in the QuAC and CoQA papers \cite{choi2018quac, reddy2019coqa} except for those with the dagger $^\dagger$. 
We randomly sampled 106 questions and manually labeled for obtaining the number with the dagger $^\dagger$.} \vspace{-0.3cm} \vspace{-0.3cm}
\label{table:question_stat}
\end{table}

\section{Comparison of Questions in QuAC and CoQA}
\label{subsec:comparison_question}

Before testing the transferability of \textsc{ExCorD} (\textsection\ref{subsec:transfer}), we compare the question distribution of QuAC to that of CoQA.
The types of questions are significantly different due to the difference in task setups.
When questions were generated in QuAC, evidence documents were soley provided to answerers, but not to questioners. 
This setup prevented questioners from referring to the evidence documents, which encouraged the questioners to ask natural and information-seeking questions.
By contrast, when creating CoQA, questioners and answerers shared the same evidence documents.

Examples of QuAC and CoQA are presented in Table \ref{table:comparison} and the categorization of question types in Table \ref{table:question_stat}.
The results are as follows:
(1) QuAC has more non-factoid questions. 
Approximately half of QuAC questions are non-factoid, whereas more than 60\% of questions in CoQA can be answered with either entities or noun phrases.
(2) ``Anything else?'' questions are more frequently observed in QuAC.
When questioners cannot find what to ask, they use ``Anything else?'' questions to seek new topics and continue the conversation.
In CoQA, questioners rarely used the ``Anything else?'' question (2.8\%) since they did not need to seek new topics.
This type of question is observed in Table \ref{table:comparison} ($q_2$ in the left side).
(3) CoQA has few unanswerable questions.
Since questioners and answerers share the evidence documents when creating CoQA, only 1.3\% of unanswerable questions are asked.
However, approximately 20\% of questions in QuAC are unanswerable.

\section{Hyperparameters}
\label{subsec:hyperparameters}

Our implementation is based on PyTorch.\footnote{\url{https://pytorch.org/}}
We implemented BERT using the Transformers library.\footnote{\url{https://github.com/huggingface}}
We implemented the T5-based QR model using the Transformers library and adopted the same QR model in the pipeline approach and \textsc{ExCorD}.
We use a single 24GB GPU (RTX TITAN) for the experiments.

We measured the F1 scores on the development set for each 4k training step, and adopted the best-performing models.
We trained QA models based on the AdamW optimizer with a learning rate of 3e-5.
We use the maximum input sequence length as 512 and the maximum answer length as 30. 
We set the maximum query length to 128 for all approaches since self-contained questions are usually longer than original questions.
We use a batch size 12 for BERT and RoBERTa in all baseline approaches.
For \textsc{ExCorD}, we set the coefficient $\lambda_1$ for QA loss for rewritten questions to 0.5. Also we search the coefficient $\lambda_2$ for consistency loss within the range of [0.7, 0.5] and the softmax temperature within the range of [1.0, 0.9] \cite{xie2019unsupervised}.

\end{document}